\documentclass{article}
\usepackage{spconf,amsmath,graphicx,hyperref}
\usepackage{cite}
\usepackage{amsmath,amssymb,amsfonts}
\usepackage{algorithmic}
\usepackage{graphicx}
\usepackage{caption}
\usepackage{textcomp}
\usepackage{multirow}
\usepackage{multicol}
\usepackage{booktabs}

\title{MoiréNet: A Compact Dual-Domain Network for Image Demoiréing}
\name{Shuwei Guo$^1$, Simin Luan$^1$, Yan Ke$^2$, Zeyd Boukhers$^3$, John See $^4$, Cong Yang$^1$$^{\dagger}$\thanks{$^{\dagger}$Cong Yang is the corresponding author.\\
Guo et al. Copyright 2026 lEEE. Personal use otthis material is permitted. Permission from IEEEmust be obtained for all other uses, includingreprinting/republishing, creating new collectiveworks, for resale or redistribution to servers or lisor reuse of any copyrighted component of thiswork. DOl will be added upon lEEE Xplorepublication}}
\address{%
\begin{minipage}{\linewidth}
\centering
$^1$ School of Future Science and Engineering, Soochow University, China \\
$^2$ Clobotics, China \quad
$^3$ Fraunhofer Institute for Applied Information Technology, Germany \\
$^4$ School of Mathematical and Computer Sciences, Heriot-Watt University (Malaysia Campus), Malaysia \hfill
\end{minipage}
}

\begin{document}
%

\maketitle
\begin{abstract}
Moiré patterns arise from spectral aliasing between display pixel lattices and camera sensor grids, manifesting as anisotropic, multi-scale artifacts that pose significant challenges for digital image demoiréing. We propose MoiréNet, a convolutional neural U-Net-based framework that synergistically integrates frequency and spatial domain features for effective artifact removal. MoiréNet introduces two key components: a Directional Frequency-Spatial Encoder (DFSE) that discerns moiré orientation via directional difference convolution, and a Frequency-Spatial Adaptive Selector (FSAS) that enables precise, feature-adaptive suppression. Extensive experiments demonstrate that MoiréNet achieves state-of-the-art performance on public and actively used datasets while being highly parameter-efficient. With only 5.513M parameters, representing a 48\% reduction compared to ESDNet-L, MoiréNet combines superior restoration quality with parameter efficiency, making it well-suited for resource-constrained applications including smartphone photography, industrial imaging, and augmented reality.
\end{abstract}
\section{Introduction}
Moiré patterns are spatial aliasing artifacts arising from spectral interference between high-frequency scene content and the camera's colour filter array (CFA)~\cite{Coarsetofine}. These patterns manifest as undesirable stripes and ripples whose frequency components often overlap with and confound those of the underlying true texture~\cite{FPANet}. Such spectral overlap necessitates sophisticated joint spatial-frequency analysis for effective removal. While some works have focused on Moiré detection~\cite{MoireDet, MoireDet+}, the primary challenge remains in the end-to-end restoration, or demoiréing, of the pristine image.

While numerous deep learning-based demoiréing methods have been proposed, they exhibit persistent limitations. Representative approaches include DMCNN~\cite{TIP2018}, using multi-scale convolutions but lacking cross-scale interaction; MopNet~\cite{MopNet}, exploiting edge and attribute cues but separating spatial and frequency processing; MBCNN~\cite{MBCNN}, adopting multi-branch implicit frequency modelling without robust cross-domain fusion; FHDe$^2$Net~\cite{FHDMI}, performing coarse-to-fine restoration but leaving residual artifacts; and ESDNet~\cite{UHDM}, introducing semantically aligned frequency cues but with high memory cost and limited explicit fusion. In general, most approaches~\cite{TIP2018, UHDM, MopNet, MBCNN, FHDMI, MDDM, WDNet, unpaired} rely on loosely coupled dual-branch or sequential designs, overlook the intrinsic directional nature of moiré fringes, and maintain large model sizes that hinder deployment on resource-constrained devices.

\begin{figure}
    \centering
    \includegraphics[height=4cm]{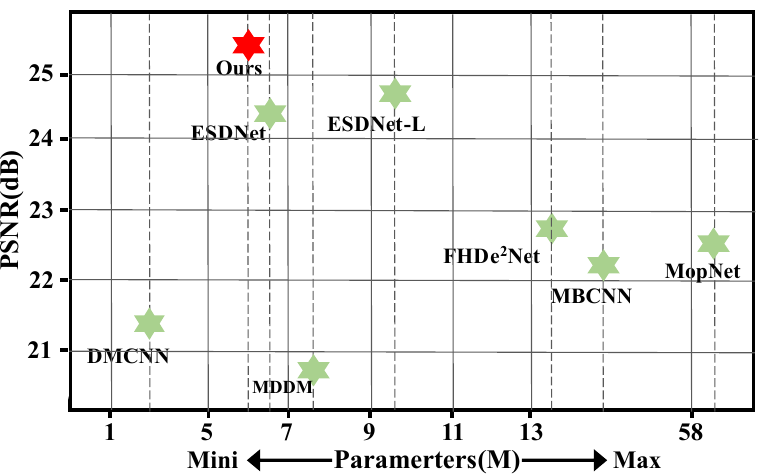}
    \caption{Comparison of PSNR vs. parameters on FHDMi~\cite{FHDMI} between MoiréNet and state-of-the-art methods.}
    \label{fig:table}
\end{figure}
\begin{figure*}[t]
  \centering
  \includegraphics[height=6.95cm]{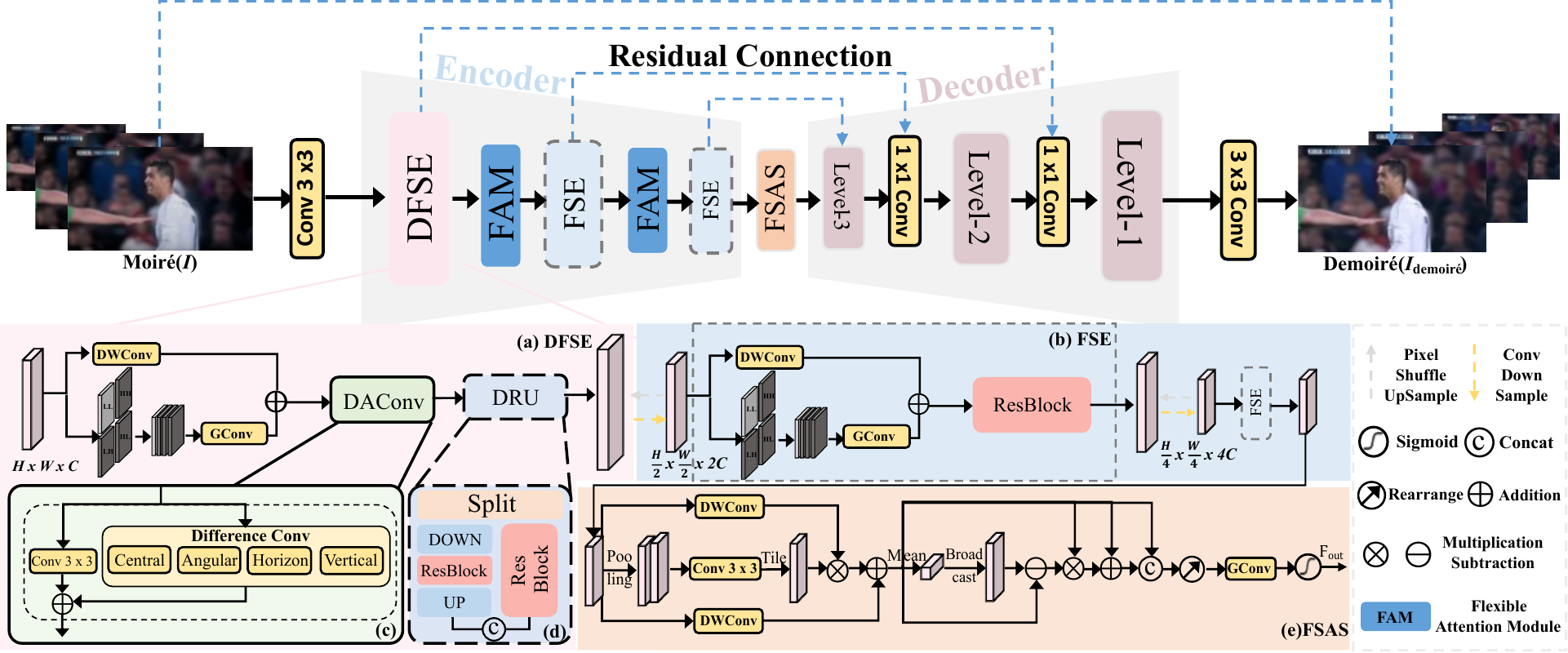}
  \caption{Overall architecture of MoiréNet.}
  \label{fig:framework}
\end{figure*}

To address these challenges, we introduce MoiréNet, a novel convolutional neural U-Net-based framework designed for efficient and effective dual-domain demoiréing. At its core, MoiréNet features two specialized encoders: a Directional Frequency-Spatial Encoder (DFSE), which leverages directional boosted convolutions to model the orientation and periodicity of Moiré patterns explicitly, and a Frequency-Spatial Encoder (FSE) for robust coarse-scale feature processing. These are complemented by a Frequency-Spatial Adaptive Selector (FSAS) to dynamically emphasize important regions and a Flexible Attention Module (FAM) for enhanced cross-stage feature fusion. Ultimately, MoiréNet establishes a new state-of-the-art, outperforming ESDNet-L~\cite{UHDM} by +0.3dB PSNR while requiring 52\% fewer parameters. Such a favorable trade-off between accuracy and parameter facilitates its deployment on edge devices for applications in smartphone photography, industrial imaging, and augmented reality.

Succinctly, the main contributions of this work are as follows: 1) We propose a dual-domain feature extraction strategy that integrates the Directional Frequency-Spatial Encoder (DFSE) and Frequency-Spatial Encoder (FSE), enabling effective modelling of directional and multi-scale characteristics of moiré patterns. 2) We design the Frequency-Spatial Adaptive Selector (FSAS) for dynamic spatial-frequency component selection, thereby enhancing artifact removal while retaining structural details. 3) We present MoiréNet, a novel compact image demoiréing framework with only 5.513M parameters (a 48\% reduction relative to ESDNet-L), which achieves state-of-the-art (SOTA) results on widely used public benchmarks, demonstrating an advantageous balance between accuracy and deployability.

\section{PROPOSED METHOD}
As illustrated in Fig.~\ref{fig:framework}, MoiréNet adopts a convolutional neural U-Net-based architecture for demoiréing. Given an input image $I \in \mathbb{R}^{H \times W \times 3}$, shallow features are first extracted and fed into the encoder, where a Directional Frequency-Spatial Encoder (DFSE) models the periodic and directional characteristics of moiré patterns, followed by Frequency-Spatial Encoders (FSE) that capture multi-scale representations. A Flexible Attention Module (FAM) aggregates features of the same resolution with residual connections to enhance information flow, while a Frequency-Spatial Adaptive Selector (FSAS) at the bottleneck adaptively emphasizes critical components. The decoder restores resolution using pixel-shuffle and transposed convolutions. The final result is obtained through residual learning, expressed as: $I_{\text{demoiré}} = I + F(I)$, where $F(I)$ is the processed feature map.

\subsection{Directional Frequency-Spatial Encoder}
The Directional Frequency-Spatial Encoder (DFSE) is designed to suppress moiré artifacts while preserving structural fidelity through joint spatial and frequency analysis. As shown in Fig.~\ref{fig:framework}(a), it comprises a Frequency-Spatial Encoder for dual-domain representation, a Dual Residual Unit for long-range dependency modelling, and a Detail Augmented Convolution for directional feature extraction. Together, these modules generate robust representations that facilitate accurate characterization and effective removal of moiré patterns.

\noindent
\textbf{Frequency-Spatial Encoder:}
Moiré patterns contain both spatial structures and high-frequency components, making single-domain processing inadequate for reliable suppression. To address this, the Frequency-Spatial Encoder (FSE) employs a dual-branch design to extract features, as shown in Fig.\ref{fig:framework} (b). Building upon~\cite{WTconv}, the spatial branch leverages depth-wise convolutions to capture local structural features, while the frequency branch applies a discrete wavelet transform (DWT) to decompose the input into four subbands (LL, LH, HL, HH), thereby isolating low-frequency structures from high-frequency artifacts. The subband features are processed by group convolution and reconstructed via inverse wavelet transform (IWT) to reproject them into the spatial domain. The outputs of the two branches are then fused by element-wise addition, and residual blocks refine fused features to enhance multi-scale representation and suppress moiré. The general operation is formulated as follows:
\begin{equation}
\hat{F} = R \Big( C_d(F) + \text{IWT} \big( C_g \big( \text{DWT}(F) \big) \big) \Big)
\label{eq:fse}
\quad .
\end{equation}
where \({F}\) denotes the input feature map. \(C_d\) and \(C_g\) represent depth-wise convolution and group convolution, respectively. \(R\) indicates residual block processing.

\noindent
\textbf{Detail Augmented Convolution:}
Moiré artifacts, arising from directional interference in fine textures and repetitive patterns, remain challenging to capture with conventional convolutions that employ isotropic receptive fields~\cite{dehazing}. To address this, we introduce the Detail-Augmented Convolution (DAC), which embeds directional priors via multiple specialized convolution branches to enhance moiré suppression.

Inspired by previous work~\cite{DEANet}, DAC extends Conv with four branches: Central Difference Convolution (CDC)~\cite{CDC} for local gradient variations, Angular Difference Convolution (ADC) for angular responses, and Horizontal/Vertical Moiré Difference Convolutions (HMDC/VMDC), which are tailored to detect orientation-specific moiré structures using gradient-inspired operators such as Scharr~\cite{Scharr}. For instance, VMDC enhances vertical moiré sensitivity by assigning positive weights to the upper row and negative weights to the lower row of a $3\times 3$ kernel.

As shown in Fig.~\ref{fig:framework} (c), the DAC integrates five parallel paths (\( k \in \{1, 2, 3, 4, 5\} \), corresponding to Conv, CDC, ADC, HMDC, VMDC), which are fused into a unified kernel via learnable coefficients \(\alpha_k\). The operation is expressed as:
\begin{equation}
\hat{F} = F * \sum_{k=1}^{5} \alpha_k W_k + \sum_{k=1}^{5} \alpha_k b_k
\quad .
\end{equation}
where \( F \) is the input feature map, \(*\) denotes convolution, and \( W_k \), \( b_k \) are the weights and biases of the \( k \)-th branch. This design preserves directional sensitivity with a parameter count comparable to standard convolution.

\noindent
\textbf{Dual Residual Unit:}
Moiré patterns exhibit fine-scale, intricate structures that are difficult to suppress with conventional residual blocks. Their limited receptive fields restrict the ability to capture long-range dependencies, which are critical for robust demoiréing. Motivated by~\cite{Inception}, we design the Dual Residual Unit (DRU), a U-Net-like block that performs dual-path feature extraction via channel splitting and resolution reduction, as shown in Fig.~\ref{fig:framework} (d). Specifically, given an input feature map, it is divided into a high-resolution branch for local detail preservation and a low-resolution branch that models global context. The latter undergoes downsampling, residual processing, and subsequent upsampling before fusion with the high-resolution path. This dual-path strategy effectively integrates local and global representations, thereby enhancing moiré suppression.
\begin{figure}
    \centering
    \includegraphics[height=4cm]{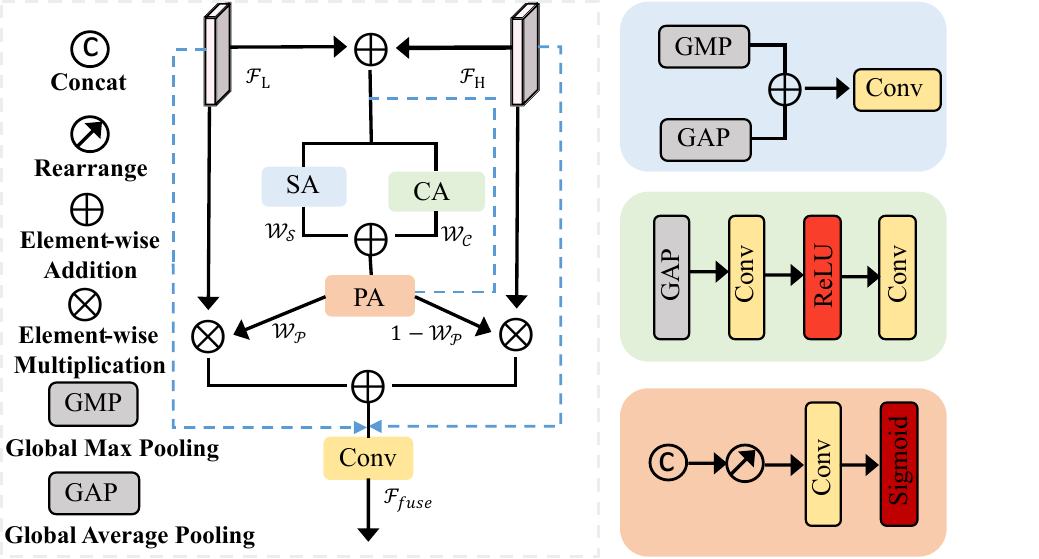}
    \caption{The diagram of the Flexible Attention Module (FAM).}
    \label{fig:FAM}
\end{figure}

\begin{table*}[t]
    \small
    \setlength{\tabcolsep}{0.06mm}
    \renewcommand{\arraystretch}{1.1} 
    \centering
    \caption{
    Quantitative comparisons in terms of PSNR and SSIM between our model and state-of-the-art methods on four datasets.
    \textbf{Bold} denotes the best and \underline{underline} the second-best results. 
    The Params indicates the number of model parameters.
    }

    \label{tab:quantitative}
    \begin{tabular}{c|c|ccccccccc|c}
        \hline
        Dataset & Metrics & Input & DMCNN~\cite{TIP2018} & MDDM~\cite{MDDM} & WDNet~\cite{WDNet} & MopNet~\cite{MopNet} & MBCNN~\cite{MBCNN} & FHDe$^2$Net~\cite{FHDMI} & ESDNet~\cite{UHDM} & ESDNet-L~\cite{UHDM} & Ours \\
        \hline
        \multirow{2}{*}{TIP2018} & PSNR$\uparrow$ & 20.30 & 26.77 & - & 28.08 & 27.75 & 30.03 & 27.78 & 29.81 & \underline{30.11} & \textbf{30.49} \\
        & SSIM$\uparrow$ & 0.7380 & 0.8719 & - & 0.9040 & 0.8950 & 0.8930 & 0.8960 & 0.9160 & \underline{0.9200} & \textbf{0.9224} \\
        \hline
        \multirow{2}{*}{LCDMoire} & PSNR$\uparrow$ & 10.44 & 35.48 & 42.49 & 29.66 & - & 44.04 & 41.40 & 44.83 & \underline{45.34} & \textbf{45.59} \\
        & SSIM$\uparrow$ & 0.5717 & 0.9785 & 0.9940 & 0.9670 & - & 0.9948 & - & \underline{0.9963} & \textbf{0.9966} & 0.9931 \\
        \hline
        \multirow{2}{*}{FHDMi} & PSNR$\uparrow$ & 17.97 & 21.54 & 20.83 & - & 22.76 & 22.31 & 22.93 & 24.50 & \underline{24.882} & \textbf{25.49} \\
        & SSIM$\uparrow$ & 0.7033 & 0.7727 & 0.7343 & - & 0.7958 & 0.8095 & 0.7885 & 0.8351 & \underline{0.8440} & \textbf{0.8506} \\
        \hline
        \multirow{2}{*}{UHDM} & PSNR$\uparrow$ & 17.12 & 19.91 & 20.09 & 20.36 & 19.49 & 21.41 & 20.34 & 22.12 & \underline{22.42} & \textbf{22.77} \\
        & SSIM$\uparrow$ & 0.5089 & 0.7575 & 0.7441 & 0.6497 & 0.7572 & 0.7932 & 0.7496 & 0.7956 & \underline{0.7985} & \textbf{0.8078} \\ \hline
        \hline
         & Params(M) & - & \textbf{1.426} & 7.637 & 3.360 & 58.565 & 14.192 & 13.571 & 5.934 & 10.623 & \underline{5.513} \\
        \hline
    \end{tabular}
\end{table*}
\begin{figure*}[t]
  \centering
  \includegraphics[width=\textwidth]{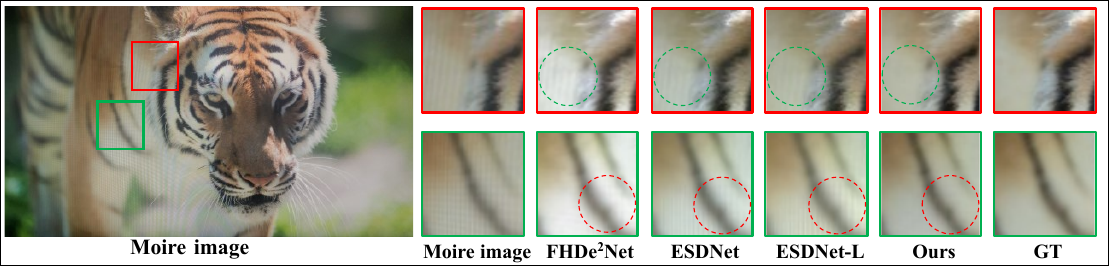}
  \caption{Visual results of different demoiréing methods on the FHDMi dataset~\cite{FHDMI}. Please zoom in for a detailed comparison.}
  \label{fig:result}
\end{figure*}

\begin{table}[t]
    \small 
    \setlength{\tabcolsep}{0.6mm}
    \renewcommand{\arraystretch}{1.15} 
    \centering
    \caption{
    Experimental Results: Performance with Crop 256.
    }
    \label{tab:256}
    \begin{tabular}{l|c|cc|c}
        \hline
        Dataset & Metrics & MBCNN~\cite{MBCNN} & ESDNet-L~\cite{UHDM} & Ours \\
        \hline
        FHDMi & PSNR\&SSIM & 21.62/0.8079 & 23.41/0.8289 & \textbf{24.03}/\textbf{0.8303} \\
        \hline
        UHDM & PSNR\&SSIM & 19.00/0.7461 & 20.41/0.7701  & \textbf{20.55}/\textbf{0.7742} \\
        \hline
    \end{tabular}
\end{table}

\begin{table}[t]
    \small
    \setlength{\tabcolsep}{1.6mm}
    \renewcommand{\arraystretch}{1.15}
    \centering
    \caption{Ablation results for the design of DFSE.}
    \label{tab:DFSE}
    \begin{tabular}{l|cc|cc}
        \hline
        Design & FSE & +DAC & +DRU   \\
        \hline
        PSNR\&SSIM  & 21.41 / 0.7896 & 22.51 / 0.8062 & \textbf{22.99 / 0.8091} \\
        \hline
    \end{tabular}
\end{table}

\begin{table}[t]
    \small 
    \renewcommand{\arraystretch}{1.05} 
    \setlength{\tabcolsep}{4mm}
    \centering
    \caption{
    Ablation Results: The designs of FSAS and FAM.
    }
    \label{tab:FSSC}
    \begin{tabular}{l|cc|cc}
        \hline
        Design & + FSAS & + FAM & PSNR &SSIM  \\
        \hline
        (a) & - &- &21.79 &0.7922  \\
        (b) & \checkmark&- & 22.26 & 0.7981  \\
        (c) & -&\checkmark &22.30 &0.8005\\
        \hline
        (d) &\checkmark &\checkmark&\textbf{22.51}&\textbf{0.8062} \\
        \hline
    \end{tabular}
\end{table}

\subsection{Frequency-Spatial Adaptive Selector}
The Frequency-Spatial Attention Selector (FSAS) enhances critical spatial-frequency components for effective demoiréing, as illustrated in Fig.\ref{fig:framework} (e). We extend the methodology of~\cite{FocalNet}, FSAS performs sequential spatial–frequency selection and pixel-level refinement to improve feature description.

Given an input feature map \( F \), FSAS first derives a spatial importance map \( F_s \) along the channel dimension via max and average pooling followed by depth-wise convolution, thereby highlighting regions susceptible to moiré. This spatial map is then element-wise multiplied with \( F \) and normalized by subtracting its mean response to emphasize interference:

\begin{equation}
F_{fs} = (F \otimes (F_s - \mathrm{mean}(F_s))) + F
\quad .
\end{equation}
where \(\otimes\) denotes element-wise multiplication and \(\mathrm{mean}\) indicates spatial averaging. On this basis, pixel-level attention is introduced by concatenating  \(F_{fs}\) with the pixel-attention response, applying grouped convolutions, and generating pixel-wise weights \( W_p \) via sigmoid activation. The final output adopts residual modulation with learnable scalars :
\begin{equation}
F_{\text{out}} = a\big(W_p \otimes F_{fs}\big) + bF
\label{eq:fsas_output}
\quad.
\end{equation}
where learnable scalars \( a \) and \( b \) are initialized to \( 0 \) and \( 1 \), respectively. This design strengthens moiré suppression while maintaining residual information, thereby improving restoration stability and overall performance.

\subsection{Flexible Attention Module}
To improve feature fusion in demoiréing, we introduce the Flexible Attention Module (FAM), which jointly integrates spatial, channel, and pixel attention mechanisms. Spatial attention~\cite{CBAM} emphasizes moiré-sensitive regions, channel attention~\cite{Squeeze} adaptively reweights feature responses, and pixel attention models fine-grained spatial–channel dependencies, as illustrated in Fig.~\ref{fig:FAM}.

FAM first combines pre-encoder features \(F_{L}\) and encoder outputs \(F_{H}\) via element-wise addition. Spatial attention~\cite{CBAM} generates a weight map $W_s$ using global pooling and convolution, while channel attention~\cite{Squeeze} produces $W_c$ via global pooling and two $1\times1$ convolutions with ReLU. These maps guide pixel attention, which concatenates them with fused features and applies grouped convolutions to obtain pixel weights \(W_p\). Finally, \(W_p\) modulates \(F_{L}\) and \(F_{H}\), and the result is projected with a \(1 \times 1\) convolution to form a residual output that preserves details and suppresses moiré artifacts.

\section{Experiments}
\subsection{Datasets and Evaluation Protocols}
We evaluate MoiréNet on four widely used datasets: UHDM~\cite{UHDM} (real 4K), FHDMi~\cite{FHDMI} (1920×1080), LCDMoire~\cite{LCDMoire} (synthetic 1024×1024), and TIP2018~\cite{TIP2018} (around 400×400). Training uses cropped patches: 768×768 for UHDM, 512×512 for FHDMi and LCDMoire, and 256×256 for TIP2018; testing uses full resolutions. The models are implemented in PyTorch and trained on NVIDIA RTX 4090 with Adam ($\beta_1{=}0.9$, $\beta_2{=}0.999$), initial learning rate $5{\times}10^{-5}$, and cyclic cosine annealing. The batch sizes are 4 for TIP2018 and 2 for others. PSNR and SSIM evaluate performance.
\subsection{Performance Evaluation}
\noindent\textbf{Overall Performance:}  
We compare MoiréNet with SOTA demoiréing methods~\cite{TIP2018, MopNet, MBCNN, FHDMI, WDNet, UHDM}. As shown in Tab.~\ref{tab:quantitative}, MoiréNet achieves the highest PSNR and SSIM with only~\textbf{5.513M} parameters. The qualitative results in Fig.~\ref{fig:result} show that it effectively removes moiré artifacts while preserving fine textures, whereas other methods leave residual artifacts or lose details.

\noindent\textbf{Robustness to Reduced Resolution:}  
To assess robustness, we trained MoiréNet, MBCNN~\cite{MBCNN}, and ESDNet-L~\cite{UHDM} on $256{\times}256$ FHDMi~\cite{FHDMI} patches. As shown in Tab.~\ref{tab:256}, MoiréNet consistently outperformed the others, demonstrating strong restoration under reduced receptive fields and effective capture of both local and global structures.
\subsection{Ablation Studies}
We perform ablation studies on FHDMi~\cite{FHDMI} using 50\% of the dataset for 250 epochs (batch size 2, learning rate $1.25{\times}10^{-5}$) to assess the contribution of each module.

\noindent
\textbf{Effectiveness of DFSE:}  
We assess DFSE by first augmenting FSE with DAC and subsequently replacing its standard residual block with DRU. The baseline FSE achieves 21.41 PSNR and 0.7896 SSIM. Incorporating DAC further improves performance, and substituting DRU yields the complete DFSE, which attains 22.99 PSNR and 0.8091 SSIM (Tab.~\ref{tab:DFSE}), demonstrating that the DFSE enhances demoiréing.

\noindent
\textbf{Effectiveness of FSAS and FAM:}  
As shown in Tab.~\ref{tab:FSSC}, integrating either FSAS or FAM individually enhances the baseline, while their joint application achieves the best results, confirming their complementary roles in moiré suppression.

\section{CONCLUSION}
In this paper, we propose MoiréNet, a novel compact image demoiréing network that unifies spatial–frequency feature extraction and models the anisotropic directionality of moiré patterns. It incorporates a Directional Frequency-Spatial Encoder to capture directional moiré structures, a Frequency-Spatial Adaptive Selector to emphasize critical dual-domain features, and a Flexible Attention Module for adaptive feature fusion. Evaluated on widely used public datasets, MoiréNet achieves state-of-the-art performance with low parameters, enabling deployment on resource-constrained edge devices.


\label{sec:refs}
\fontsize{10pt}{10pt}\selectfont 
\bibliographystyle{IEEEbib}
\bibliography{refs}

\begin{thebibliography}{10}

\bibitem{Coarsetofine}
Ce~Wang, Bin He, Shengsen Wu, Renjie Wan, Boxin Shi, and Ling-Yu Duan,
\newblock ``Coarse-to-fine disentangling demoir{\'e}ing framework for recaptured screen images,''
\newblock {\em IEEE Transactions on Pattern Analysis and Machine Intelligence}, vol. 45, no. 8, pp. 9439--9453, 2023.

\bibitem{FPANet}
Gyeongrok Oh, Sungjune Kim, Heon Gu, Sang~Ho Yoon, Jinkyu Kim, and Sangpil Kim,
\newblock ``Fpanet: Frequency-based video demoireing using frame-level post alignment,''
\newblock {\em Neural Networks}, vol. 184, pp. 107021, 2025.

\bibitem{MoireDet}
Cong Yang, Zhenyu Yang, Yan Ke, Tao Chen, Marcin Grzegorzek, and John See,
\newblock ``Doing more with moir{\'e} pattern detection in digital photos,''
\newblock {\em IEEE Transactions on Image Processing}, vol. 32, pp. 694--708, 2023.

\bibitem{MoireDet+}
Zhuocheng Li, Xizhu Shen, Simin Luan, Shuwei Guo, Zeyd Boukhers, Wei Sui, Yuyi Wang, and Cong Yang,
\newblock ``Moir{\'e} pattern detection: Stability and efficiency with evaluated loss function,''
\newblock in {\em International Conference on PatItern Recognition}, 2024, pp. 34--49.

\bibitem{TIP2018}
Yujing Sun, Yizhou Yu, and Wenping Wang,
\newblock ``Moir{\'e} photo restoration using multiresolution convolutional neural networks,''
\newblock {\em IEEE Transactions on Image Processing}, vol. 27, no. 8, pp. 4160--4172, 2018.

\bibitem{MopNet}
Bin He, Ce~Wang, Boxin Shi, and Ling-Yu Duan,
\newblock ``Mop moire patterns using mopnet,''
\newblock in {\em IEEE International Conference on Computer Vision}, 2019, pp. 2424--2432.

\bibitem{MBCNN}
Bolun Zheng, Shanxin Yuan, Chenggang Yan, Xiang Tian, Jiyong Zhang, Yaoqi Sun, Lin Liu, Ale{\v{s}} Leonardis, and Gregory Slabaugh,
\newblock ``Learning frequency domain priors for image demoireing,''
\newblock {\em IEEE Transactions on Pattern Analysis and Machine Intelligence}, vol. 44, no. 11, pp. 7705--7717, 2021.

\bibitem{FHDMI}
Bin He, Ce~Wang, Boxin Shi, and Ling-Yu Duan,
\newblock ``Fhde2net: Full high definition demoireing network,''
\newblock in {\em European Conference on Computer Vision}, 2020, pp. 713--729.

\bibitem{UHDM}
Xin Yu, Peng Dai, Wenbo Li, Lan Ma, Jiajun Shen, Jia Li, and Xiaojuan Qi,
\newblock ``Towards efficient and scale-robust ultra-high-definition image demoir{\'e}ing,''
\newblock {\em European Conference on Computer Vision}, 2022.

\bibitem{MDDM}
Xi~Cheng, Zhenyong Fu, and Jian Yang,
\newblock ``Multi-scale dynamic feature encoding network for image demoir{\'e}ing,''
\newblock in {\em IEEE International Conference on Computer Vision Workshop}, 2019, pp. 3486--3493.

\bibitem{WDNet}
Lin Liu, Jianzhuang Liu, Shanxin Yuan, Gregory Slabaugh, Ale{\v{s}} Leonardis, Wengang Zhou, and Qi~Tian,
\newblock ``Wavelet-based dual-branch network for image demoir{\'e}ing,''
\newblock in {\em European Conference on Computer Vision}, 2020, pp. 86--102.

\bibitem{unpaired}
Yunshan Zhong, Yuyao Zhou, Yuxin Zhang, Fei Chao, and Rongrong Ji,
\newblock ``Learning image demoir{\'e}ing from unpaired real data,''
\newblock in {\em AAAI Conference on Artificial Intelligence}, 2024, vol.~38, pp. 7623--7631.

\bibitem{WTconv}
Shahaf~E Finder, Roy Amoyal, Eran Treister, and Oren Freifeld,
\newblock ``Wavelet convolutions for large receptive fields,''
\newblock in {\em European Conference on Computer Vision}, 2024, pp. 363--380.

\bibitem{dehazing}
Haiyan Wu, Yanyun Qu, Shaohui Lin, Jian Zhou, Ruizhi Qiao, Zhizhong Zhang, Yuan Xie, and Lizhuang Ma,
\newblock ``Contrastive learning for compact single image dehazing,''
\newblock in {\em IEEE Conference on Computer Vision and Pattern Recognition}, 2021, pp. 10551--10560.

\bibitem{DEANet}
Zixuan Chen, Zewei He, and Zhe-Ming Lu,
\newblock ``Dea-net: Single image dehazing based on detail-enhanced convolution and content-guided attention,''
\newblock {\em IEEE Transactions on Image Processing}, vol. 33, pp. 1002--1015, 2024.

\bibitem{CDC}
Zitong Yu, Chenxu Zhao, Zezheng Wang, Yunxiao Qin, Zhuo Su, Xiaobai Li, Feng Zhou, and Guoying Zhao,
\newblock ``Searching central difference convolutional networks for face anti-spoofing,''
\newblock in {\em IEEE conference on computer vision and pattern recognition}, 2020, pp. 5295--5305.

\bibitem{Scharr}
Hanno Scharr,
\newblock ``Optimal operators in digital image processing,''
\newblock 2000.

\bibitem{Inception}
Chenyang Si, Weihao Yu, Pan Zhou, Yichen Zhou, Xinchao Wang, and Shuicheng Yan,
\newblock ``Inception transformer,''
\newblock {\em Advances in Neural Information Processing Systems}, vol. 35, pp. 23495--23509, 2022.

\bibitem{FocalNet}
Yuning Cui, Wenqi Ren, Xiaochun Cao, and Alois Knoll,
\newblock ``Focal network for image restoration,''
\newblock in {\em IEEE International Conference on Computer Vision}, 2023, pp. 13001--13011.

\bibitem{CBAM}
Sanghyun Woo, Jongchan Park, Joon-Young Lee, and In~So Kweon,
\newblock ``Cbam: Convolutional block attention module,''
\newblock in {\em European Conference on Computer Vision}, 2018, pp. 3--19.

\bibitem{Squeeze}
Jie Hu, Li~Shen, and Gang Sun,
\newblock ``Squeeze-and-excitation networks,''
\newblock in {\em IEEE Conference on Computer Vision and Pattern Recognition}, 2018, pp. 7132--7141.

\bibitem{LCDMoire}
Shanxin Yuan, Radu Timofte, Gregory Slabaugh, Ale{\v{s}} Leonardis, Bolun Zheng, Xin Ye, Xiang Tian, Yaowu Chen, Xi~Cheng, Zhenyong Fu, et~al.,
\newblock ``Aim 2019 challenge on image demoireing: Methods and results,''
\newblock in {\em IEEE International Conference on Computer Vision Workshop}, 2019, pp. 3534--3545.

\end{thebibliography}
\end{document}